\documentclass[letterpaper]{article} 
\usepackage{arxiv} 
\usepackage{times}  
\usepackage{helvet}  
\usepackage{courier}  
\usepackage[hyphens]{url} 
\usepackage{graphicx}
\usepackage{natbib} 
\usepackage{caption} 
\usepackage{algorithm}
\usepackage{algorithmic}
\usepackage{amsmath}
\usepackage{amssymb}
\usepackage{bm}
\usepackage{booktabs}
\usepackage{multirow}
\usepackage{array}
\usepackage{tabularx}
\usepackage{xcolor}
\usepackage{newfloat}
\usepackage{listings}
\usepackage{hyperref}
\usepackage{xcolor}
\DeclareCaptionStyle{ruled}{labelfont=normalfont,labelsep=colon,strut=off} 
\floatstyle{ruled}
\newfloat{listing}{tb}{lst}{}
\floatname{listing}{Listing}
\definecolor{CustomBlue}{RGB}{0, 0, 200}
\hypersetup{
    colorlinks=true,
    linkcolor=CustomBlue,
    citecolor=CustomBlue,
    filecolor=magenta,
    urlcolor=CustomBlue
}

\title{Simultaneous Training of First- and Second-Order Optimizers in Population-Based Reinforcement Learning}
\author{
    Felix Pfeiffer\textsuperscript{\rm 1}
    Shahram Eivazi\textsuperscript{\rm 1, \rm 2}
}
\affiliations{
    \textsuperscript{\rm 1}University of Tübingen\\
    \textsuperscript{\rm 2}Festo SE \& Co. KG\\
    felix.pfeiffer@protonmail.com
    
}

\begin{document}

\maketitle

\begin{abstract}
The tuning of hyperparameters in reinforcement learning (RL) is critical, as these parameters significantly impact an agent's performance and learning efficiency. Dynamic adjustment of hyperparameters during the training process can significantly enhance both the performance and stability of learning. Population-based training (PBT) provides a method to achieve this by continuously tuning hyperparameters throughout the training. This ongoing adjustment enables models to adapt to different learning stages, resulting in faster convergence and overall improved performance. In this paper, we propose an enhancement to PBT by simultaneously utilizing both first- and second-order optimizers within a single population. We conducted a series of experiments using the TD3 algorithm across various MuJoCo environments. Our results, for the first time, empirically demonstrate the potential of incorporating second-order optimizers within PBT-based RL. Specifically, the combination of the K-FAC optimizer with Adam led to up to a 10\% improvement in overall performance compared to PBT using only Adam. Additionally, in environments where Adam occasionally fails, such as the Swimmer environment, the mixed population with K-FAC exhibited more reliable learning outcomes, offering a significant advantage in training stability without a substantial increase in computational time.

\end{abstract}

\section{Introduction}
The majority of RL algorithms are sensitive to hyperparameter settings, initialization, and the stochastic nature of the environment. Regularization, robust training procedures, and stable optimization methods are necessary to enhance the overall performance of an RL agent \citep{pmlr-v130-zhang21n}. In response to the time needed for training of RL, there has been a growing interest in the ability to simultaneously explore multiple regions of the hyperparameter space using PBT \citep{jaderberg2017population,paul2019fast,wu2020accelerating,zhao2023maximum}. In this paper, we also focus on PBT and its ability to simultaneously explore multiple regions of the hyperparameter space for training of RL agents. However, we do not include techniques used in automated machine learning (Auto ML) community \citep{feurer2019hyperparameter}.

Unlike traditional training methods that focus on optimizing a single model, PBT maintains a population of evolving models that are updated either sequentially (computationally efficient but time-consuming) or in parallel across the populations as fast as a model training \citep{flajolet2022fast}. Although these techniques themselves benefit from a large number of samples, PBT algorithms suffer from their reliance on heuristics for hyperparameter tuning, which can lead to underperformance without extensive computational resources and often result in suboptimal outcomes. As such, many studies reported in PBT literature focus on the efficient training of populations. For example, \citet{parker2020provably} introduced Population-Based Bandits (PB2), a novel algorithm that maintains the PBT framework but incorporates a probabilistic model to guide hyperparameter search efficiently. In a recent study, \citet{grinsztajn2023winner} introduced Poppy, a population-based RL method that trains a set of complementary policies to maximize performance across a distribution of problem instances. Rather than explicitly enforcing diversity, Poppy uses a novel training objective that encourages agents to specialize in different subsets of the problem distribution.

More related to our work \citet{cui2018evolutionary} introduced the Evolutionary Stochastic Gradient Descent (ESGD) framework, which integrates traditional Stochastic Gradient Descent (SGD) with gradient-free evolutionary algorithms to enhance the optimization of deep neural networks. By alternating between SGD and evolutionary steps, ESGD improves the fitness of a population of candidate solutions, ensuring that the best-performing solution is preserved. This approach leverages the strengths of both optimization techniques, resulting in more effective training across a range of applications, including speech and image recognition, as well as language modeling.

On the other hand, while the use of different optimizers for efficient neural network training is well-studied \citep{choi2019empirical}, there appears to be a significant gap in the PBT literature, where the exploration of various optimizers remains largely unexplored. Although there are approaches of mixing different optimization strategies, such as in \citet{landro2020mixing}, which merges two first-order optimizers directly. In this paper, we aim to investigate the impact of utilizing different optimizers within the PBT framework. Drawing inspiration from recent findings by \citet{tatzel2022late}, which emphasize the advantages of second-order methods in the later stages of neural network training, our study seeks to leverage the benefits of diverse optimization strategies throughout the entire training process.

Second-order methods in the RL domain have been validated in various settings \citep{kamanchi2021generalized,gebotys2022understanding,salehkaleybar2022momentum} and algorithms like Natural Policy Gradient (NPG) or Trust Region Policy Optimization (TRPO). Although these methods are computationally more intensive, techniques \citep{martens2015optimizing} and libraries \citep{dangel2019backpack,gao_pytorch-kfac} have been developed to extract second-order information from a backward pass more easily. We develop and evaluate a PBT-based framework that allows agents with different optimizers to be used in one population. This approach not only capitalizes on the efficiency of first-order methods in early training stages but also utilizes the precision of second-order optimizer as the population converges. By allowing these diverse agents to coexist and compete within the population, we hypothesize that our method will achieve more robust and efficient optimization.

\section{Preliminaries}
Optimization in neural networks involves adjusting the network's parameters (weights and biases) to minimize the loss function, which measures the discrepancy between the predicted and true values. The training data is represented as $\{\bm{x}^{(i)}, \bm{y}^{(i)}\in\mathbb{R}^I\times\mathbb{R}^C\}_{i\in\mathbb{D}}$, where $\mathbb{D}=\{1,...,N\}, (\bm{x}^{(i)}, \bm{y}^{(i)})\overset{\text{iid}}{\sim} P_\text{data}$ are i.i.d. samples from the data distribution. The model $f\colon\mathbb{R}^D\times\mathbb{R}^I\to\mathbb{R}^C$ makes prediction $\hat{\bm{y}}=f(\bm{\theta}, \bm{x})$ given parameters $\bm{\theta}\in\mathbb{R}^D$ and input $\bm{x}\in\mathbb{R}^I$. The loss function $\ell\colon\mathbb{R}^C\times\mathbb{R}^C\to\mathbb{R}$ compares the prediction $\hat{\bm{y}}$ to the true label $\bm{y}$. The expected risk is defined as $\mathcal{L}_{P_{\text{data}}}(\bm{\theta})=\mathbb{E}_{(\bm{x},\bm{y}\sim P_\text{data})}[\ell (f(\bm{\theta},\bm{x}),\bm{y})]$. The goal is to find $\bm{\theta}_\ast$ that minimized the empirical risk:

\begin{align}
    \begin{split}
        \bm{\theta}_\ast &= \arg \min_{\bm{\theta}\in\mathbb{R}^D}\mathcal{L}_\mathbb{D}(\bm{\theta}) \\
        &\quad\quad\text{with} \\
        \mathcal{L}_\mathbb{D}(\bm{\theta}) &= \frac{1}{\vert\mathbb{D}\vert}\sum_{i\in\mathbb{D}}\ell (f(\bm{\theta},\bm{x}^{(i)}),\bm{y}^{(i)})
    \end{split}
\end{align}

For our study, we used the Adam optimization algorithm, an adaptive gradient method introduced by \citet{kingma2014adam}. Adam combines the advantages of AdaGrad and RMSprop, adjusting the learning rate for each parameter based on the first and second moments of the gradients. 

\subsection{Second Order Optimization}
The Newton step is a central concept in second-order optimization methods. The Newton method incorporates second-order information through the Hessian matrix, which captures the curvature of the loss landscape.
The Hessian, denoted as $\bm{H} = \nabla^2 \mathcal{L}_{P_{\text{data}}}(\bm{\theta})$, is a square matrix of second-order partial derivatives of the loss function. 
To derive the Newton step, we start by considering a quadratic Taylor approximation of the loss function around $\bm{\theta}_t$:

\begin{align}
    \begin{split}
        \mathcal{L}_{P_{\text{data}}}(\bm{\theta}) \approx q(\bm{\theta}) &:= \mathcal{L}_{P_{\text{data}}}(\bm{\theta}_t) \\
        &+ (\bm{\theta} - \bm{\theta}_t)^\top \bm{g} \\
        &+ \frac{1}{2} (\bm{\theta} - \bm{\theta}_t)^\top \bm{H} (\bm{\theta} - \bm{\theta}_t)
    \end{split}
\end{align}

Where $\bm{g} := \nabla \mathcal{L}_{P_{\text{data}}}(\bm{\theta}_t)$ and $\bm{H} := \nabla^2 \mathcal{L}_{P_{\text{data}}}(\bm{\theta}_t)\succ0$. The Hessian is assumed to be positive definite. To get the next iterate $\bm{\theta}_{t+1}$, we set $\nabla q = \bm{g} + \bm{H}(\bm{\theta}-\bm{\theta}_t) \overset{!}{=} 0$. We obtain the Newton step:

\begin{align}
    \bm{\theta}_{t+1} = \bm{\theta}_t-\bm{H}^{-1}\bm{g}
\end{align}

The Newton method can lead to faster convergence near the optimum due to its potential for quadratic convergence, offering a significant speed advantage over first-order methods in certain scenarios. However, computing and inverting the Hessian matrix is computationally expensive and can be impractical for large neural networks, with a complexity of $\mathcal{O}(n^2)$ for storage, $\mathcal{O}(n^3)$ for inversion, and $\mathcal{O}(n^2)$ for computing the Hessian, where $n$ is the number of parameters of the neural network.

\subsection{Dealing with Non-Convex Objective Functions: The Generalized Gauss-Newton Matrix}
When optimizing non-convex objective functions, the Hessian matrix may have negative eigenvalues, making it indefinite and potentially leading to non-minimizing steps. This is a common challenge in deep learning due to the highly non-convex loss landscape. The Generalized Gauss-Newton (GGN) matrix offers a solution by providing a positive semi-definite approximation of the Hessian \cite{schraudolph2002fast, martens2020new}. The GGN matrix is derived by decomposing the Hessian based on the model's output, specifically through the mapping $\bm{\theta} \mapsto f(\bm{\theta}, \bm{x}) \mapsto \ell(f(\bm{\theta}, \bm{x}), \bm{y})$. This allows for expressing the gradient of the loss function with respect to the parameters $\bm{\theta}$.

\begin{align}
    \nabla_{\bm{\theta}} \ell(f(\bm{\theta}, \bm{x}), \bm{y}) = \bm{J}^\top \nabla_f \ell(f(\bm{\theta}, \bm{x}), \bm{y})
\end{align}

Here, $\bm{J}\in\mathbb{R}^{C\times D}$ is the Jacobian matrix of the model's output $f(\bm{\theta}, \bm{x})$ with respect to the parameters $\bm{\theta}$.

The Hessian matrix $\bm{H}_f\in\mathbb{R}^{C\times C}$ can then be decomposed into two terms. The first term involves the Jacobian and the Hessian of the loss with respect to the model's output. The second term in the Hessian decomposition arises from the chain rule for second derivatives. This term accounts for the curvature introduced by the model parameters themselves, where $\nabla^2_{\bm{\theta}} [f(\bm{x}, \bm{\theta})]_c$ is the second derivative of the $c$-th output of the model with respect to the parameters $\bm{\theta}$. $\nabla_f \ell(f(\bm{\theta}, \bm{x}), \bm{y})]_c$ is the gradient of the loss with respect to the $c$-th output of the model:

\begin{align}
    \begin{split}
        &\hspace{25mm}\nabla^2_{\bm{\theta}} \ell(f(\bm{\theta}, \bm{x}), \bm{y}) = \\
        &\bm{J}^\top \bm{H}_f \bm{J} + \sum_{c=1}^C \nabla^2_{\bm{\theta}} [f(\bm{x}, \bm{\theta})]_c \cdot [\nabla_f \ell(f(\bm{\theta}, \bm{x}), \bm{y})]_c
    \end{split}
\end{align}

The GGN $\bm{G}$ simplifies this by neglecting the second term and focusing only on the first term:

\begin{align}
    \bm{G} = \mathbb{E}_{(\bm{x},\bm{y}) \sim P_{\text{data}(\bm{x},\bm{y})}} [\bm{J}^\top \bm{H}_f \bm{J}]
\end{align}

This expectation over the data distribution ensures that $\bm{G}$ captures the average curvature of the loss function with respect to the parameters. Since $\bm{J}^\top \bm{H}_f \bm{J}$ represents a quadratic from, the matrix is positive semi-definite.

\subsection{Damping and the Trust-Region Problem}
When the objective function is convex, the quadratic model approximating the objective function can be arbitrarily bad. To address this issue, the concept of a trust region is introduced \cite{more1983computing}. The idea is to limit the parameter updates to a region where the quadratic model $q$ is a good approximation of the actual objective function. This is achieved by restricting the parameter updates to lie within a certain radius $r$ around the current parameter $\bm{\theta}_t$. Mathematically, the trust-region problem can be formulated as:

\begin{align}
    \bm{\theta}_{t+1} = \arg \min_{\bm{\theta}} q(\bm{\theta}) \quad \text{such that} \quad \|\bm{\theta}_{t+1} - \bm{\theta}_t\|\leq r
\end{align}

Combining the trust-region problem with the Newton step gives us a modified version of Newton's method given by:

\begin{align}\label{eq:newton-step-with-damping}
    \bm{\theta}_{t+1} = \bm{\theta}_t - (\bm{H} + \delta \bm{I})^{-1} \bm{g} \quad \text{with} \quad \delta \geq 0
\end{align}

Here, $\delta$ is the damping parameter. The damping term $\delta \bm{I}$, where $\bm{I}$ is the identity matrix, is added to the Hessian to control the step size. Damping can be seen as interpolation between a first-order and a second-order optimization step.

\subsection{Diagonal Gauss-Newton Second Order Optimizer}
The Diagonal Generalized Gauss-Newton (Diag. GGN) optimizer is an approach designed to simplify the computation and inversion of the Gauss-Newton matrix by focusing only on its diagonal elements. The full GGN captures the curvature information comprehensively. However, considering only the diagonal elements, the Diag. GGN optimizer approximates the curvature more coarsely. This approximation assumes that the off-diagonal elements are less significant, which may not always be true. The update formula for the parameters is similar to the update formula of the Newton step (\ref{eq:newton-step-with-damping}). It is given by:

\begin{align}\label{eq:diag-ggn-optimizer}
    \bm{\theta}_{t+1} = \bm{\theta}_t - \alpha(\bm{G}(\bm{\theta}_t) + \delta \bm{I})^{-1} \bm{g}(\bm{\theta}_t)
\end{align}

 Where $\alpha$ is the step size, $\bm{G}$ is the diagonal of the GGN, and $\bm{g}$ is the gradient.

\subsection{Kronecker-Factored Approximate Curvature Optimizer}
The Kronecker-Factored Approximate Curvature (K-FAC) optimizer, introduced by \citet{martens2015optimizing}, is a second-order method that efficiently approximates the Fisher Information Matrix (FIM). The FIM measures the amount of information an observable variable carries about an unknown parameter and is closely related to the curvature of the likelihood function with respect to model parameters. K-FAC simplifies the inversion of the FIM by approximating it with a block-diagonal matrix, where each block corresponds to parameters of a specific layer or group of layers, significantly reducing computational complexity. For a linear layer, each block of the FIM is structured as follows:

\begin{align}
    \hat{\bm{F}}_i = \mathbb{E}[a_i a_i^\top \otimes b_i b_i^\top] \approx \mathbb{E}[a_i a_i^\top] \otimes \mathbb{E}[b_i b_i^\top] =: \bm{A}_i \otimes \bm{B}_i
\end{align}

Here, $a_i$ represents the input activations, and $b_i$ represents the gradients with respect to the pre-activations. By approximating the FIM as a Kronecker product of two smaller matrices $\bm{A}_i$ and $\bm{B}_i$, we can efficiently compute the inverse of each block:

\begin{align}
    \hat{\bm{F}}_i^{-1} = (\bm{A}_i \otimes \bm{B}_i)^{-1} = \bm{A}_i^{-1} \otimes \bm{B}_i^{-1}
\end{align}

Mathematically, the parameter update rule of K-FAC is given by:

\begin{align}
    \bm{\theta}_{t+1} = \bm{\theta}_t - \alpha (\bm{A}_i^{-1} \otimes \bm{B}_i^{-1}) \nabla_{\bm{\theta}}\mathcal{L}(\bm{\theta}_t)
\end{align}

\section{Experiments}
In this paper, all experiments were performed using the Twin Delayed Deep Deterministic Policy Gradient (TD3) algorithm based on the widely used CleanRL library \cite{huang2022cleanrl}. TD3 is the successor of the Deep Deterministic Policy Gradient (DDPG) algorithm \cite{lillicrap2015continuous}, which combines the ideas from Deep Q-Network (DQN) \cite{mnih2015human} and Deterministic Policy Gradient (DPG) \cite{silver2014deterministic}.

\subsection{Simulation Tasks}
We used the MuJoCo environments V4 from the Gymnasium library, including HalfCheetah, Hopper, Humanoid, Swimmer, and Walker2d. These scenarios test RL algorithms in continuous control tasks, aiming to maximize the agent's traversal distance, with each episode limited to 1000 steps.

We ensured consistency of our experiments by using the same hyperparameters as those used in the RL Baselines3 Zoo, as well as performance comparison. Each environment underwent ten independent training runs for single-agent experiments. For the PBT experiments, we trained five populations. Consistent with the RL Baselines3 Zoo bwnchmark, we trained the agents for one million steps. After the training phase, we performed 150,000 evaluation steps to measure the performance.

\subsection{Second-Order Optimizer}
The Diag. GGN optimizer was implemented using the BackPACK library \cite{dangel2019backpack}. For the K-FAC optimizer we used the PyTorch implementation by \citet{gao_pytorch-kfac}.

\begin{figure}[h!]
  \centering
  \includegraphics[width=\columnwidth]{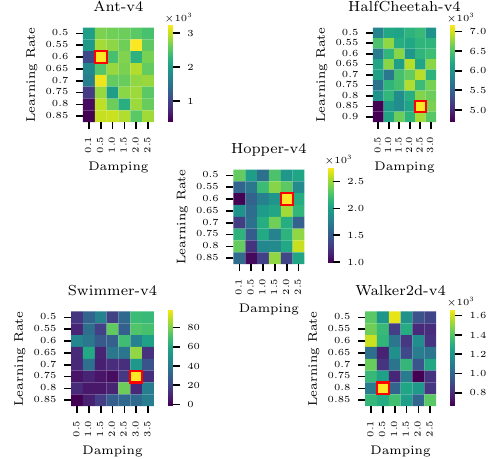}
  \caption{Grid search for Diag. GGN across multiple environments, displaying mean rewards for various hyperparameter settings.}
  \label{fig:grid_search_ggn}
\end{figure}

Due to the absence of established hyperparameters for second-order optimizers in the context of TD3, we conducted a thorough grid search. This process focused on learning rate and damping parameters. For each combination, we trained ten models over 500,000 steps and evaluated them after training for 150,000 steps to obtain the final performance. This systematic approach allowed us to identify effective hyperparameter configurations for the Diag. GGN and K-FAC optimizers across various MuJoCo environments. Figure \ref{fig:grid_search_ggn} presents the grid search for the Diag. GGN optimizer.

\begin{figure}[h!]
  \centering
  \includegraphics[width=\columnwidth]{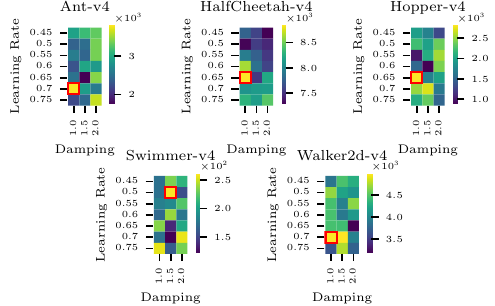}
  \caption{Grid search for K-FAC across multiple environments, displaying mean rewards for various hyperparameter settings.}
  \label{fig:grid_search_kfac}
\end{figure}

Similarly, Figure \ref{fig:grid_search_kfac} shows the grid search for the K-FAC optimizer. During our experiments, we encountered consistent failures with the K-FAC optimizer when the damping parameter was set below 1.0. This issue arose because the Cholesky factorization could not be computed, as the inputs were not positive-definite.

\subsection{Population Based Training}
Our implementation and hyperparameters selection are based on the PBT approach by \citet{jaderberg2017population}, where the bottom 20\% of agents copy hyperparameters from the top 20\%, with random perturbations by factors of 0.8 or 1.2.

When replacing an agent with another using the same optimizer, we transfer network weights, replay buffer, and hyperparameters, then perturb the hyperparameters (including batch size, learning rates, and, for the second-order optimizer, additionally, the damping parameter). For agents with different optimizers, only network weights and the replay buffer is transferred, while the receiving agent's hyperparameters are perturbed.

\subsection{Population Size}
In this study, we built on findings from previous research highlighting the importance of population size in PBT. Notably, \citet{bai2024generalized} demonstrated that significant performance gains could be achieved with as few as four agents, with diminishing returns beyond eight to sixteen agents. Similarly, \citet{shahid2024scaling} found that a population size of eight optimally balances exploration and computational effort.

\addtolength{\tabcolsep}{-3.5pt}    
\begin{table}[b!]
        \begin{tabular}{lccc}
        \toprule
        \multirow{2}{*}{Environment} & \multicolumn{3}{c}{Population Size} \\
        \cmidrule(lr){2-4}
        & 4 & 8 & 16 \\
        \midrule
        Ant-v4         & $3669 \pm 1173$ & $4703 \pm 939$  & $5420 \pm 1070$ \\
        HalfCheetah-v4 & $9755 \pm 826$  & $10655 \pm 948$ & $11158 \pm 459$ \\
        Hopper-v4      & $3521 \pm 107$  & $3452 \pm 155$  & $3461 \pm 191$ \\
        Swimmer-v4     & $236 \pm 153$   & $299 \pm 159$   & $363 \pm 1$ \\
        Walker2d-v4    & $4629 \pm 579$  & $4833 \pm 548$  & $5092 \pm 360$ \\
        \bottomrule
        \end{tabular}
    \caption{Influence of the population sizes on the performance.}
    \label{tab:reward_different_population_sizes}
\end{table}
\addtolength{\tabcolsep}{+3,5pt} 

Based on these insights, we conducted experiments with populations of four, eight, and sixteen agents, using a perturbation interval of 10,000 steps. Results over one million training steps, evaluated across 150,000 steps with five different random seeds, showed that increasing population size generally improved rewards, especially from four to eight agents. Especially in the Ant and HalfCheetah environments (See \ref{tab:reward_different_population_sizes}). During our study, we observed that agents in the Swimmer environment tend to exhibit a binary outcome: they either fail completely, receive minimal rewards, or succeed in achieving a consistent reward of approximately 360. This dichotomy in performance is the reason for the high standard deviation observed in the Swimmer results throughout all our results. Despite improved rewards with larger populations, the benefits of expanding to sixteen agents were marginal, leading us to select a population size of eight.

\section{Results}
\begin{figure}[h!]
  \centering
  \includegraphics[width=\columnwidth]{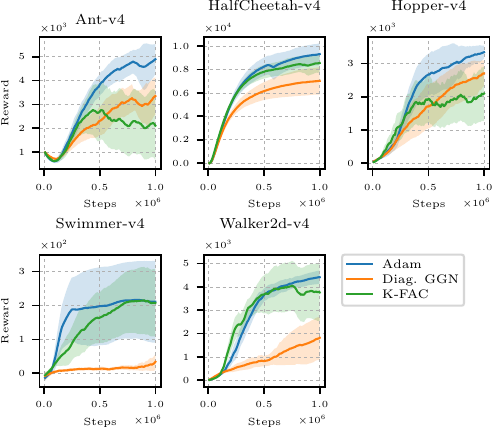}
  \caption{Mean reward during training for single agents using different optimizer.}
  \label{fig:single_agents_diff_optims_reward_during_training}
\end{figure}

In the initial experiments, we independently compared each optimizer. Table \ref{tab:single_agent_diff_optim} displays the rewards at the end of training. The rewards during training are shown in Figure \ref{fig:single_agents_diff_optims_reward_during_training}. Adam consistently outperformed the second-order methods in all environments.

\addtolength{\tabcolsep}{-3.5pt}    
\begin{table}[h!]
    \begin{tabular}{lccc}
    \toprule
    Environment & Adam & Diag. GGN & K-FAC \\
    \midrule
    Ant-v4         & $5145 \pm 719$  & $3802 \pm 956$  & $1548 \pm 2645$ \\
    HalfCheetah-v4 & $9495 \pm 1204$ & $7119 \pm 1575$ & $8696 \pm 1081$ \\
    Hopper-v4      & $3447 \pm 141$  & $2737 \pm 609$  & $2302 \pm 1441$ \\
    Swimmer-v4     & $213 \pm 155$   & $71 \pm 114$    & $204 \pm 129$ \\
    Walker2d-v4    & $4499 \pm 382$  & $2043 \pm 1174$ & $3582 \pm 1630$ \\
    \bottomrule
    \end{tabular}
    \caption{Performance comparison of different optimizer across various environments}
    \label{tab:single_agent_diff_optim}
\end{table}
\addtolength{\tabcolsep}{+3.5pt}    

\subsection{First and Second-Order Optimizers in one Population}
\begin{figure}[h!]
  \centering
  \includegraphics[width=\columnwidth]{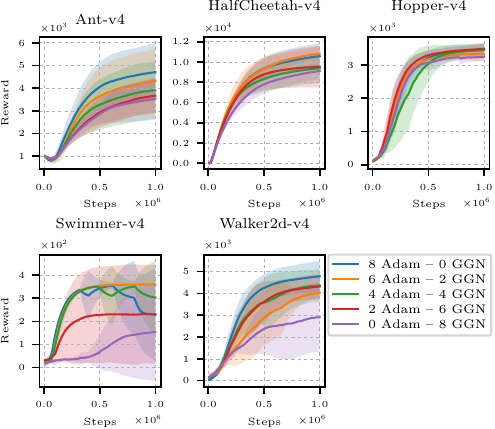}
  \caption{Training of Adam and Diag. GGN in one population.}
  \label{fig:pbt_adam_diag_ggn_reward_during_training}
\end{figure}

Table \ref{tab:reward_stats_adam_gnn} shows the reward at the end of training when using Adam and Diag. GGN optimizers in one population. Figure \ref{fig:pbt_adam_diag_ggn_reward_during_training} shows the results that were achieved during training. In general, adding Diag. GGN optimizer to the population improved the performance in more than half of the environments. Specially we observed that in two Diag. GGN settings, the Swimmer environment no longer fails, and therefore, we can see significant improvement compared to PBT training using only Adam optimizers (See Table \ref{tab:reward_stats_improve_adam_gnn}).

\begin{table*}[tb!]
    \centering
    \begin{tabularx}{\textwidth}{l*{5}{>{\centering\arraybackslash}X}}
    \toprule
        \multirow{2}{*}{Environment} & \multicolumn{5}{c}{Num. Adam Agents -- Num. Diag. GGN Agents} \\
        \cmidrule(lr){2-6}
        & 8 -- 0 & 6 -- 2 & 4 -- 4 & 2 -- 6 & 0 -- 8 \\
        \midrule
        Ant-v4         & $\bm{4703 \pm 939}$  & $4304 \pm 1029$ & $3892 \pm 855$  & $3670 \pm 555$  & $3513 \pm 652$ \\
        HalfCheetah-v4 & $10655 \pm 948$ & $\bm{10881 \pm 563}$ & $9472 \pm 1016$ & $9597 \pm 1518$ & $9197 \pm 605$ \\
        Hopper-v4      & $3452 \pm 155$  & $3328 \pm 180$  & $3478 \pm 112$  & $\bm{3493 \pm 128}$  & $3261 \pm 76$ \\
        Swimmer-v4     & $229 \pm 159$   & $\bm{361 \pm 1}$     & $295 \pm 130$   & $234 \pm 153$   & $154 \pm 153$ \\
        Walker2d-v4    & $\bm{4833 \pm 548}$  & $4109 \pm 182$  & $4381 \pm 535$  & $4385 \pm 450$  & $2882 \pm 1240$ \\
        \bottomrule
    \end{tabularx}
    \caption{Summary of reward statistics: Mean and standard deviation for mixed poulations of Adam and Diag. GGN agents}
    \label{tab:reward_stats_adam_gnn}

    \vspace{1.0em}
    
    \centering
    \begin{tabularx}{\textwidth}{l*{5}{>{\centering\arraybackslash}X}}
        \toprule
        \multirow{2}{*}{Environment} & \multicolumn{5}{c}{Num. Adam Agents -- Num. K-FAC Agents} \\
        \cmidrule(lr){2-6}
        & 8 -- 0 & 6 -- 2 & 4 -- 4 & 2 -- 6 & 0 -- 8 \\
        \midrule
        Ant-v4         & $4703 \pm 939$  & $4327 \pm 674$  & $4650 \pm 886$  & $\bm{5129 \pm 630}$  & $4364 \pm 791$ \\
        HalfCheetah-v4 & $10655 \pm 948$ & $\bm{11106 \pm 784}$ & $10598 \pm 582$ & $11014 \pm 534$ & $9071 \pm 561$ \\
        Hopper-v4      & $3452 \pm 155$  & $\bm{3724 \pm 38}$   & $3695 \pm 62$   & $3569 \pm 80$   & $3617 \pm 61$ \\
        Swimmer-v4     & $229 \pm 159$   & $\bm{361 \pm 2}$     & $\bm{361 \pm 1}$     & $359 \pm 2$     & $350 \pm 13$ \\
        Walker2d-v4    & $4833 \pm 548$  & $5192 \pm 529$  & $4588 \pm 680$  & $5245 \pm 374$  & $\bm{5265 \pm 333}$ \\
        \bottomrule
    \end{tabularx}
    \caption{Summary of reward statistics: Mean and standard deviation for mixed poulations of Adam and K-FAC agents}
    \label{tab:reward_stats_adam_k-fac}
\end{table*}

\begin{figure}[h!]
  \centering
  \includegraphics[width=\columnwidth]{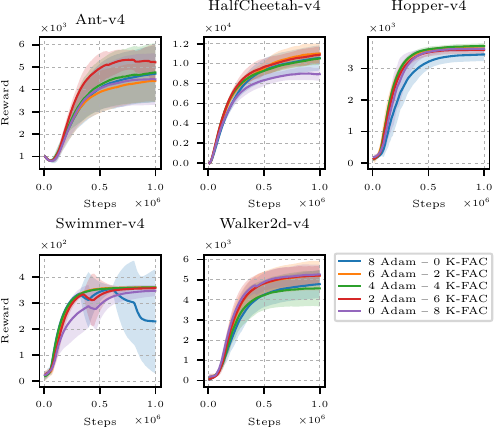}
  \caption{Training of Adam and K-FAC in one population.}
  \label{fig:pbt_adam_kfac_reward_during_training}
\end{figure}

\begin{table}[h!]
    \centering
    \begin{tabular}{lcc}
    \toprule
    \multirow{2}{*}{Environment} & \multicolumn{2}{c}{Improvement in $\% \Delta$ over} \\
    \cmidrule(lr){2-3}
    & Single Agent & PBT Adam \\
    \midrule
    Ant-v4         & $-16$ & $-9$ \\
    HalfCheetah-v4 & $+15$ & $+2$\\
    Hopper-v4      & $+1$ & $+1$\\
    Swimmer-v4     & $+69$ & $+58$ \\
    Walker2d-v4    & $-3$ & $-9$ \\
    \bottomrule
    \end{tabular}
    \caption{Improvement in percent of mixed population with Adam and Diag. GGN agents over single agents and PBT using only Adam.}
    \label{tab:reward_stats_improve_adam_gnn}
\end{table}

Table \ref{tab:reward_stats_adam_k-fac} presents the reward at the end of training for using Adam and K-FAC optimizer in one population. Figure \ref{fig:pbt_adam_kfac_reward_during_training} illustrates the results obtained during the training process. In all cases, using K-FAC improved the performance. We also observed that the reward increases faster in all environments besides Swimmer. Similar to adding the Diag. GGN optimizer setting, the K-FAC optimizer also helps the agents in the Swimmer environment to no longer fail (See \ref{tab:reward_stats_improve_adam_K-fac}). Based on our experiments, the PBT of the Ant environment seemed challenging (especially with a small population) By adding K-FAC, we were able to achieve the same rewards of single-agent training. Additionally, earlier, we observed that the increase in population size did not improve the performance in the Hopper environment. Adding K-FAC to the population for the first time increased the performance in this environment significantly.

\begin{table}[h!]
    \centering
    \begin{tabular}{lccc}
    \toprule
    \multirow{2}{*}{Environment} & \multicolumn{2}{c}{Improvement in $\% \Delta$ over} \\
    \cmidrule(lr){2-3}
    & Single Agent & PBT Adam \\
    \midrule
    Ant-v4         & $\pm0$ & $+10$ \\
    HalfCheetah-v4 & $+16$ & $+8$\\
    Hopper-v4      & $+8$ & $+8$\\
    Swimmer-v4     & $+69$ & $+58$ \\
    Walker2d-v4    & $+17$ & $+9$ \\
    \bottomrule
    \end{tabular}
    \caption{Improvement in percent of mixed population with Adam and K-FAC agents over single agents and PBT using only Adam.}
    \label{tab:reward_stats_improve_adam_K-fac}
\end{table}

\begin{table*}[h!]
    \centering

    \begin{tabularx}{\textwidth}{l*{5}{>{\centering\arraybackslash}X}}
        \toprule
        \multirow{2}{*}{Environment} & \multicolumn{5}{c}{Num. Adam Agents -- Num. Diag. GGN Agents} \\
        \cmidrule(lr){2-6}
        & 8 -- 0 & 6 -- 2 & 4 -- 4 & 2 -- 6 & 0 -- 8 \\
        \midrule
        Ant-v4         & $\bm{4703 \pm 939}$  & $3236 \pm 691$   & $3309 \pm 373$  & $3263 \pm 304$  & $2688 \pm 427$ \\
        HalfCheetah-v4 & $\bm{10655 \pm 948}$ & $10402 \pm 1884$ & $10451 \pm 875$ & $8947 \pm 1050$ & $6329 \pm 951$ \\
        Hopper-v4      & $3452 \pm 155$  & $3358 \pm 156$   & $\bm{3459 \pm 152}$  & $3359 \pm 138$  & $2511 \pm 859$ \\
        Swimmer-v4     & $229 \pm 159$   & $\bm{362 \pm 2}$      & $227 \pm 158$   & $170 \pm 150$   & $141 \pm 125$ \\
        Walker2d-v4    & $\bm{4833 \pm 548}$  & $4525 \pm 515$   & $3964 \pm 473$  & $4426 \pm 603$  & $1956 \pm 838$  \\
        \bottomrule
    \end{tabularx}
    \caption{Summary of reward statistics: Mean and standard deviation for mixed populations of Adam and Diag. GGN agents. Step count adjusted.}
    \label{tab:gnn_step_count}
    
    \vspace{1.0em}
    
    \begin{tabularx}{\textwidth}{l*{5}{>{\centering\arraybackslash}X}}
        \toprule
        \multirow{2}{*}{Environment} & \multicolumn{5}{c}{Num. Adam Agents -- Num. K-FAC Agents} \\
        \cmidrule(lr){2-6}
        & 8 -- 0 & 6 -- 2 & 4 -- 4 & 2 -- 6 & 0 -- 8 \\
        \midrule
        Ant-v4         & $\bm{4703 \pm 939}$  & $3739 \pm 997$   & $3375 \pm 235$  & $3592 \pm 173$  & $3335 \pm 198$ \\
        HalfCheetah-v4 & $\bm{10655 \pm 948}$ & $10439 \pm 1250$ & $10243 \pm 594$ & $10035 \pm 675$ & $3921 \pm 1857$ \\
        Hopper-v4      & $3452 \pm 155$  & $3640 \pm 64$    & $\bm{3617 \pm 18}$   & $3611 \pm 88$   & $3523 \pm 122$ \\
        Swimmer-v4     & $229 \pm 159$   & $\bm{360 \pm 2}$      & $358 \pm 4$     & $359 \pm 3$     & $171 \pm 148$ \\
        Walker2d-v4    & $4833 \pm 548$  & $4870 \pm 552$   & $4163 \pm 430$  & $\bm{5000 \pm 331}$  & $3155 \pm 1192$ \\
        \bottomrule
    \end{tabularx}
    \caption{Summary of reward statistics: Mean and standard deviation for mixed populations of Adam and K-FAC agents. Step count adjusted.}
    \label{tab:k-fac_step_count}
\end{table*}

\section{Balancing Wall-Clock Runtime Across Optimizers}

We conducted a measurement of the wall-clock runtime for three optimizers (Adam, diag. GGN, and K-FAC) on our compute cluster, using an Intel XEON CPU E5-2650 v4 and Nvidia GeForce GTX 1080 Ti. Table \ref{tab:wall_clock_runtime} shows the wall-clock runtime in secons for one PBT interval, consisting of 10,000 gradient steps, with the parallel execution of four workers. 

\begin{table}[H]
    \centering
    \begin{tabular}{lccc}
        \toprule
        Environment & Adam [s] & Diag. GGN [s] & K-FAC [s] \\
        \midrule
        Ant-v4 & $102 \pm 11$ & $182 \pm 5$ & $365 \pm 43$ \\
        HalfCheetah-v4 & $96 \pm 12$ & $180 \pm 4$ & $382 \pm 45$ \\
        Hopper-v4 & $105 \pm 8$ & $180 \pm 3$ & $415 \pm 42$ \\
        Swimmer-v4 & $91 \pm 8$ & $177 \pm 5$ & $377 \pm 53$ \\
        Walker2d-v4 & $97 \pm 8$ & $185 \pm 6$ & $363 \pm 63$ \\
        \bottomrule
    \end{tabular}
    \caption{Adam, Diag. GGN and K-FAC wall-clock runtime in seconds for 10,000 gradient steps.}
    \label{tab:wall_clock_runtime}
\end{table}

We further evaluate the effectiveness of our mixed population approach by normalizing runtime across first and second-order optimizers through adjusted step counts (i.e., optimizer gradient steps). Table \ref{tab:gnn_step_count} shows performance differences in populations with Adam and Diag. GNN. In this scenario, Adam executed 10,000 gradient steps, whereas Diag. GNN completed 5,000 steps. Similarly, Table \ref{tab:k-fac_step_count} compares the populations using Adam and K-FAC, where Adam again performed 10,000 gradient steps, but K-FAC executed only 3,000 steps. As anticipated, the overall performance of our approach experienced a decline. However, certain population settings demonstrated marginally better results in the Hopper task when utilizing Diag. GNN and Adam. Moreover, the Swimmer task was reliably learned. When employing Adam and K-FAC, the performance remained $5\%$ higher in the Hopper and $3\%$ higher in the Walker2d environment. Additionally, the Swimmer task continued to be learned reliably again.

\section{Discussion}
In this study, with the lessons learned from the second order optimization research \citep{kamanchi2021generalized,tatzel2022late,salehkaleybar2022momentum,gebotys2022understanding} for the first time, to our knowledge, here we propose to use both first and second order optimizer simultaneously for efficient training of PBT-based RL. While second-order methods are less popular in deep learning settings due to computation cost as well as implementation complexity, investigating second-order optimization techniques in PBT-based RL is interesting because of their potential to enhance the efficiency and effectiveness of learning in complex environments. 

In this paper, we first show that a well-tuned Adam optimizer consistently outperformed the second-order methods in all environments when trained as a single agent. Second, overall training second-order methods with Adam in one population not only improve the performance but also help agent to avoid failing in environments like Swimmer. We provide empirical evidence that using Adam and K-FAC optimizer in one population demonstrates significant improvement. 

We further demonstrate that even under the same runtime constraints, the use of both first-order and second-order optimizer simultaneously often results in better performance than using only the Adam optimizer in the population. This suggests that the diversity of optimization strategies within a population can lead to more robust learning outcomes. For further work we propose including additional second-order or different first-order methods into one population. Expanding this diversity to include different RL algorithms could also enhance performance and stability.

\bibliography{bib}

\end{document}